\documentclass[draftclsnofoot,onecolumn]{IEEEtran}
\IEEEoverridecommandlockouts
\usepackage{cite}
\usepackage{amsmath,amssymb,amsfonts}
\usepackage{algorithmic}
\usepackage{graphicx}
\usepackage{textcomp}
\usepackage{xcolor}
\usepackage{booktabs}
\usepackage{multirow}
\usepackage{array}
\usepackage{newtxmath}
\def\BibTeX{{\rm B\kern-.05em{\sc i\kern-.025em b}\kern-.08em
    T\kern-.1667em\lower.7ex\hbox{E}\kern-.125emX}}
\begin{document}

\title{Periodic Graph-Enhanced Multivariate Time Series Anomaly Detector}

\author{\IEEEauthorblockN{Jia Li, Shiyu Long, Ye Yuan$^*$\thanks{* corresponding author. This research is supported in part by the National Natural Science Foundation of China under grants 62372385.}}

\IEEEauthorblockA{\textit{College of Computer and Information Science} \\
\textit{Southwest University}\\
Chongqing, China \\
lj112023321001551@email.swu.edu.cn, isabella333@email.swu.edu.cn, yuanyekl@swu.edu.cn}
}


\maketitle

\begin{abstract}
Multivariate time series (MTS) anomaly detection commonly encounters in various domains like finance, healthcare, and industrial monitoring. However, existing MTS anomaly detection methods are mostly defined on the static graph structure, which fails to perform an accurate representation of complex spatio-temporal correlations in MTS. To address this issue, this study proposes a \underline{P}eriodic \underline{G}raph-Enhanced \underline{M}ultivariate Time Series \underline{A}nomaly Detector (PGMA) with the following two-fold ideas: a) designing a periodic time-slot allocation strategy based Fast Fourier Transform (FFT), which enables the graph structure to reflect dynamic changes in MTS; b) utilizing graph neural network and temporal extension convolution to accurate extract the complex spatio-temporal correlations from the reconstructed periodic graphs. Experiments on four real datasets from real applications demonstrate that the proposed PGMA outperforms state-of-the-art models in MTS anomaly detection. 
\end{abstract}


\section{Introduction}

With the rapid development of sensor networks, multivariate time series (MTS) data have grown explosively across fields such as finance, healthcare, transport, industry, and meteorology\cite{1,a,b,aa,bb}. MTS are usually collected from multiple sensors or observation points, where each sensor records a one-dimensional sequence\cite{2,c,d,cc,dd}, and their combination reflects the system’s state and trends. Compared with single time series, MTS are more complex as each dimension evolves over time while interacting with others\cite{3,e,ee}. However, effectively modeling the interactions and dynamic evolution of these time series poses both opportunities and challenges for anomaly detection.


Traditional anomaly detection methods rely on density\cite{f,ff}, similarity\cite{g,h,gg,hh}, and posterior probability\cite{i,j,ii,jj} of time series\cite{41}. Density- and similarity-based methods detect outliers by measuring distances, e.g., KNN\cite{k,kk}, LOF\cite{l,ll}, while posterior probability-based methods use probabilistic models, e.g., GMM\cite{m,mm}, Bayesian Networks\cite{n,o,nn,oo}. For multivariate time series, some approaches treat them as combinations of univariate series and merge results after separate detection, but this ignores spatio-temporal relationships\cite{p,q,pp,qq}.

With the rapid development of deep learning, its strong nonlinear fitting ability has driven significant progress in multivariate time series anomaly detection\cite{r,s,t,rr,ss,tt}. Many methods reconstruct time series and detect anomalies by leveraging sensor feature relationships through graph neural networks (GNN)\cite{15}. However, challenges remain:

1) Sensor data exhibit complex nonlinear topological relationships that are difficult to model. Fully connected graphs may introduce noise, while sparse graphs based on cosine similarity treat the structure as fixed, ignoring temporal variations in dependencies. Thus, learning dynamic and nonlinear sensor relationships remains a key issue.

2) Time series have intricate temporal correlations. Temporal Convolutional Networks (TCNs)\cite{19} can capture long dependencies and improve anomaly detection, but prior studies often rely on point-level inputs lacking semantic context. While TCNs preserve sequence order via causal and dilated convolutions, their performance is sensitive to kernel size and depth. Enhancing TCNs with multi-scale filters and dilation remains an important research direction.

To mitigate the above limitations, we propose a novel PGMA (Periodic Graph-Enhanced Multivariate Time Series Anomaly Detector), which adapts to changes in relationships between variables over different time periods by dynamically reconfiguring the graph structure. At the same time, an extended temporal convolutional module is introduced to enhance the model's ability to learn temporal features. In this way, we are able to capture the dynamic evolution in multivariate time series data more comprehensively and thus improve the accuracy of anomaly detection. The main contributions of this paper include:
\begin{itemize}
\item We propose a periodic graph reconstruction spatio-temporal convolutional network for anomaly detection, allowing the graph structure to capture dynamic changes in time series.
\item We design a graph construction method based on periodic patterns and introduce an extended temporal convolution module to better fuse temporal and spatial information.
\item Extensive experiments verify the effectiveness and superiority of the proposed method in various scenarios.
\end{itemize}

\section{Related Works}

\begin{figure*}[t!]
\centerline{\includegraphics[width=17cm]{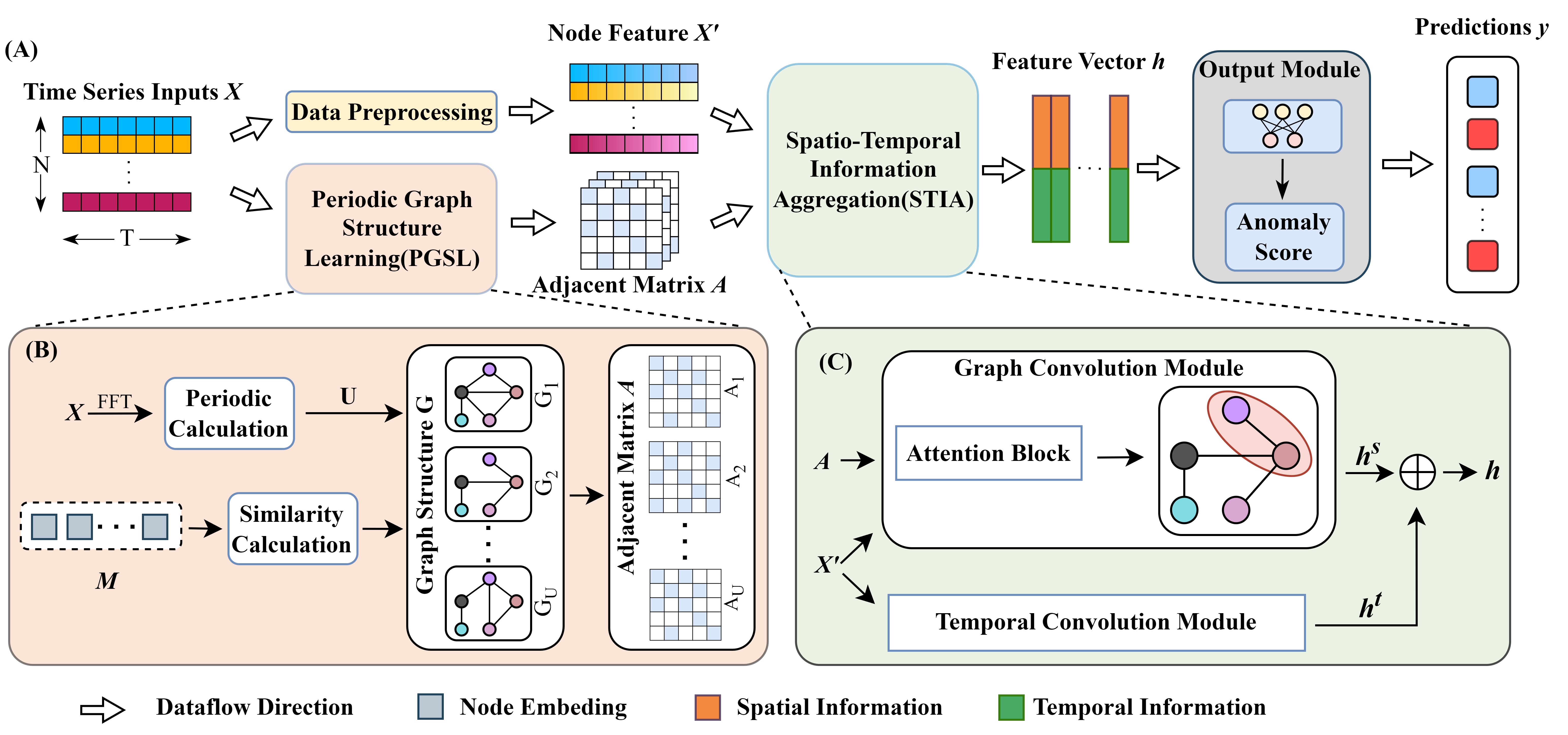}}
\caption{Framework of PGMA, show in (A), consisting of Data Preprocessing, PGSL, STIA, and Output Module three components. The inputs are first mapped to the potential space through a linear layer, while the inputs are transformed in the frequency domain using the FFT, which is used to come up with the periodic features and generate the corresponding graph structure for each periodic pattern. Then, the temporal convolution module and the graph convolution module act on the features to capture the temporal dependencies and spatial dependencies, respectively. The output module projects hidden features to the desired dimension to get the final results. Subfigures (B) and (C) denote the specific process of PGSL and STIA.}
\label{fig:model}
\end{figure*}

\subsection{Traditional Anomaly Detection Methods}

Traditional anomaly detection methods often rely on density, similarity, or posterior probability, such as KNN and Bayesian networks\cite{u,v,w,uu,vv,ww}. With the rise of deep learning, more advanced models have been proposed. RNNs\cite{23} capture temporal correlations but struggle with long sequences. VAE-LSTM\cite{24} combines VAE for short-term and LSTM for long-term features, while OmniAnomaly\cite{25} integrates GRU and VAE to model temporal dependence and stochasticity. Anomaly Transformer\cite{26} captures local and global correlations from periodic patterns, TimesNet\cite{27} transforms periodic information via FFT for CNN-based feature extraction, PatchTST\cite{39} models inter-patch dependencies, and FRNet\cite{40} rotates frequency components to capture periodic patterns. Despite these advances, many methods underutilize spatial correlations, limiting performance in complex scenarios.

\subsection{Graph Neural Network in Multivariate Time Series}

In recent years, GNNs have been widely applied to multivariate time series anomaly detection for their ability to model spatial correlations among variables\cite{x,y,z,xx,yy,zz}. Since explicit graph structures are often absent, various methods have been proposed: MTGNN\cite{29} learns inter-variable dependencies with generalized graph structures; GTS\cite{30} optimizes discrete graphs end-to-end; GAT\cite{31} introduces attention for local and global information; GCN\cite{32}, MAD-SGCN\cite{33}, and GReLeN\cite{34} enhance spatial–temporal feature extraction. To address missing neighborhood information, GDN\cite{35} employs attention with top-k selection, while DyGraphAD\cite{36} models dynamic edge weights. Although these approaches improve spatial modeling, most overlook the evolving nature of graph structures, limiting their ability to fully capture spatio-temporal dynamics.

\section{Proposed PGMA}

\subsection{Problem Description}\label{AA}
Multivariate time series consist of \emph{N} features and \emph{T} timestamps, defined as $X\mathrm{\in}\mathbb{R}^{N\times T}$, with each feature originating from a sensor. Same as other methods, the training dataset is defined as $X_{train}=[x_{train}^{(1)},\cdots,x_{train}^{(T_{train})}]$, where $x_{\mathrm{train}}^{(\mathbf{t})}\in\mathbb{R}^{N}$. Similar to other unsupervised anomaly detection methods, the training data does not include any anomaly samples. The testing dataset is defined as $X_{test}=[x_{test}^{(1)},\cdots,x_{test}^{(T_{test})}]^{}$. The main task of the anomaly detection model is to generate a set of binary labels indicating whether an anomaly exists at each timestamp in the testing set. The model achieves this by outputting a binary vector $y^{\mathrm{t}}={model(x^{\mathrm{t}})}$, where $y^{\mathrm{t}}\in\mathbb{R}^{k}$ and $y^{\mathrm{t}}\in\{0,1\}$. Here, $y^{\mathrm{t}}=1$ indicates that an anomaly is present at timestamp.

\subsection{Overview of the Framework}

The overall structure of PGMA is shown in Fig. 1, which consists of four modules: Data Preprocessing, Periodic Graph Structure Learning (PGLS), Spatio-Temporal Information Aggregation (STIA), and Output. First, the input data are normalized and passed through a linear layer to extract temporal features. Then, PGLS constructs a dynamic graph by capturing periodic dependencies among sensors, reflecting both spatial relationships and temporal variations. STIA further aggregates spatial features through graph convolution and fuses them with temporal features to enhance spatio-temporal representation. Finally, the Output module processes fused features through a fully connected layer, computes anomaly scores, and generates binary anomaly labels with adaptive thresholds.

\subsection{Periodic Graph Structure Learning}

A key goal of our framework is to learn sensor relationships across different periodic patterns through graph structures. Unlike fixed graphs, the periodic design captures both periodic features and dynamic dependencies in multivariate time series. By segmenting the series into periodic intervals, the model gains richer contextual information and better understands feature correlations. To achieve this, we first identify periodic patterns using Fast Fourier Transform (FFT) in the following way:
\begin{equation}
L=\mathrm{Avg}(\mathrm{Amp}(\mathrm{FFT}(X_{_{1\mathrm{D}}})))
\end{equation}
\begin{equation}
p=\left\lceil\frac{T}{f_L}\right\rceil,
\end{equation}
where, FFT(·) and Amp(·) denote the FFT and the calculation of amplitude values. $L$ represents the calculated amplitude, which is averaged from $N$ dimensions and each frequency by Avg(·), $p$ is the periodic length.

After obtaining the periodic patterns, we construct the graph structure in each periodic segment. Methodologically, we first initialize a matrix $M{\in}\mathbb{R}^{N\times d}$, including the features of each node, and subsequently compute the relationships between nodes based on this matrix.
\begin{equation}
E=\frac{MM^T}{\|M\|\cdot\|M\|},
\end{equation}
\begin{equation}
A_{ji} =
\begin{cases}
1, & \text{if } j \in \mathrm{TopK}\left( \{ E_{ki} \mid k \in [0, N-1] \} \right) \\
0, & \text{otherwise}
\end{cases}
.
\end{equation}

That is, we first compute \emph{E}, the similarity between the embedding vectors of each node. In order to save computational cost and avoid overly dense graphs, we choose the top \emph{k} such similarities: here TopK denotes the index of the top \emph{k} values in its input. The value of \emph{k} can be chosen by the user according to the desired sparsity level.

\subsection{Spatio-Temporal Information Aggregation}
This section consists of a graph convolution module to extract spatial dependencies between nodes and a temporal convolution module to capture their temporal information, with detailed settings shown below.


\textbf{Graph Convolution Module.} To better capture complex node relationships, we introduce a graph attention–based feature extractor, which integrates each node with its neighbors by weighting and fusing their information through graph convolution. It is shown as follows:
\begin{equation}
h_i^s=\operatorname{Re}\operatorname{LU}(\alpha_{i,i}Wx_i^{\prime}+\sum_{j\in N(i)}\alpha_{i,j}Wx_j^{\prime}),
\end{equation}
where $h_i^s$ is spatial feature vector of node $i$, $x_{i}^{\prime}\in\mathbb{R}^{d}$ is the input feature of node $i$, $N(i)=\{j|A_{ji}>0\}$ is the set of neighbors of node $i$ obtained from the learned adjacency matrix A, $W{\in}\mathbb{R}^{d^{\prime}\times d}$ is a learnable weight matrix which applies a shared linear transformation to every node, and the attention coefficients $\alpha_{ij}$ are computed as:
\begin{equation}
V=WM,
\end{equation}
\begin{equation}
e_{ij}=\mathrm{LeakyRelu}(a^\top[\nu_i\parallel\nu_j]),
\end{equation}
\begin{equation}
\alpha_{i,j}=\frac{\exp(e_{ij})}{\sum_{k\in N(i)\cup\{i\}}\exp(e_{ik})},
\end{equation}
where $\parallel$ denotes the concatenation operation, $a$ represents trainable parameters, $\nu_i$ and $\nu_j$ indicate the embedding of node $i$ and node $j$ after the shared linear transformation.

\begin{figure}[t!]
\centerline{\includegraphics[width=7.5cm]{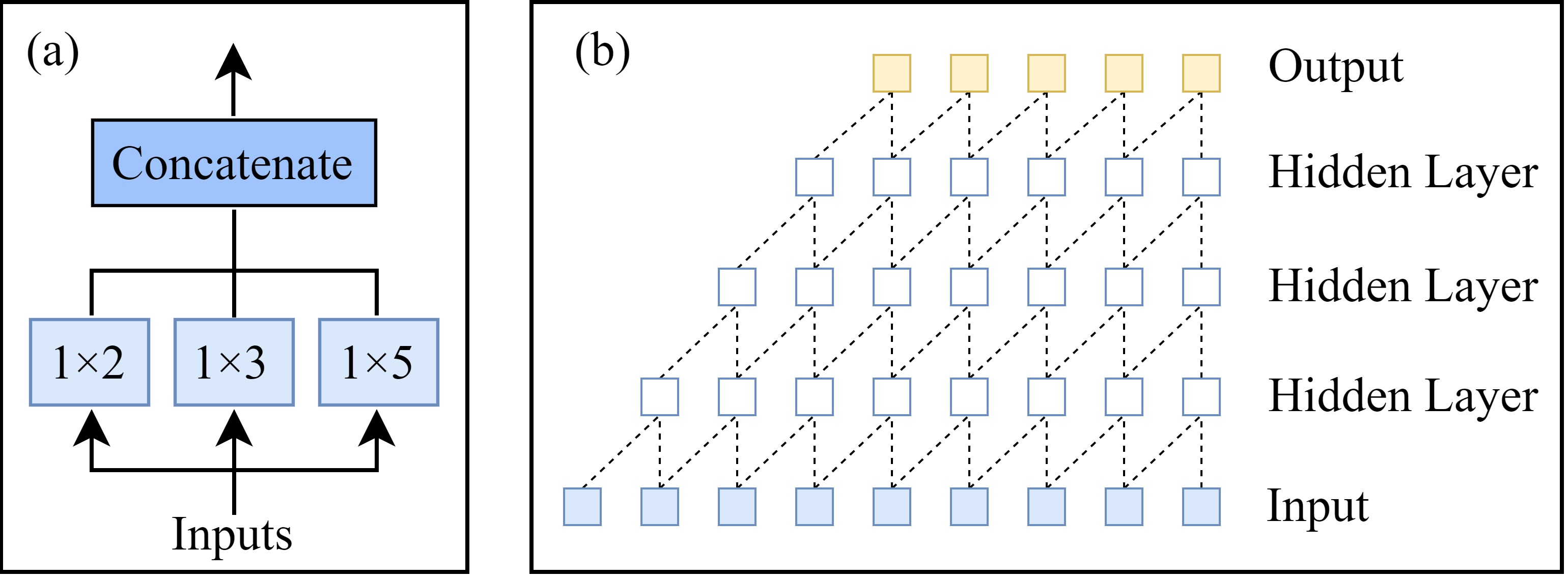}}
\caption{The Temporal Convolution Module(left) and the specific process of 1×2 filter(right).}
\label{fig:TCN}
\end{figure}


\textbf{Temporal Convolution Module.} To extract temporal information, we employ extended 1D convolutional filters combining multi-scale kernels and dilated convolution, as shown in Fig. 2. This design captures both short- and long-term temporal patterns, addressing the challenge of kernel size selection. Inspired by the inception strategy in image processing, we use filters of sizes 1×1, 1×3, and 1×5 to flexibly cover common periodic patterns (e.g., 7, 12, 24, 60). For instance, a periodic pattern of 7 can be represented by sequentially applying a 1×5 and then a 1×3 filter.

Specifically, given a 1D sequence of inputs $x\in\mathbb{R}^L$, the filter consists of $f_{1\times2}\in\mathbb{R}^{2}$, $f_{1\times3}\in\mathbb{R}^{3}$, $f_{1\times5}\in\mathbb{R}^{5}$, which are of the form:
\begin{equation}
h^t=\operatorname{Re}\operatorname{LU}([(x*f_{1\times2}) \parallel (x*f_{1\times3}) \parallel (x*f_{1\times5})]),
\end{equation}
where $h^t$ is temporal feature vector, the outputs of the three filters are truncated to the same length according to the length of the largest filter and concatenated in the channel dimension, and the dilated convolution expressed in terms of  $x*f_{1\times c}$ is defined as:
\begin{equation}
x*f_{1\times c}\left(t\right)=\sum_{s=0}^{c-1}f_{1\times c}\left(s\right)x(t-q\times s),
\end{equation}
where \emph{q} is the dilation factor.

We concatenate the outputs of the graph convolution and temporal convolution modules to combine their feature information. This approach preserves the independent features of each module while enabling subsequent layers to learn their relationships. Specifically, for each node $i$, the graph convolution and temporal convolution modules output $h_i^s$ and $h_i^t$, which are concatenated into a new feature vector $a$ as follows:
\begin{equation}
h_i=[h_i^t\parallel h_i^s],
\end{equation}
where $h_i$ is fused feature vector from graph convolution module and temporal convolution module.

\subsection{Output Module}

After the above aggregation of spatio-temporal information, a representation of all \emph{N} nodes, i.e., $\{h_{1},\cdots,h_{N}\}$ are obtain. Then, each $h_i$ is adopted to generate the prediction $\hat{X}$ at the time slot \emph{t} as:
\begin{equation}
\hat{X}=MLP(LayerNorm(h)).
\end{equation}
where \textit{MLP} is a multilayer perceptron and \textit{LayerNorm} denotes a layer normalization.
Subsequently, the \( L_2 \) loss between the predicted values $\hat{X}$ and the observed values \( X \) is employed as the primary objective function:
\begin{equation}
L_{2}=\frac1N\sum_{i=1}^N(X_i-\hat{X}_i)^2.
\end{equation}
Then, anomaly detection is performed based on the graph deviation scoring method, which is first proposed in GDN. First, the anomaly scores of each sensor at individual time points are calculated, and then the anomaly scores at each time point are combined to obtain a unified anomaly score. For the $i$-th sensor, the normalized anomaly score $ano_i(t)$ at time slot $t$ is calculated as follows:
\begin{equation}
Err_i^t=\left|X_i(t)-\hat{X}_i(t)\right|,
\end{equation}
\begin{equation}
ano_i(t)=\frac{Err_i^t-\tilde{\mu}_i}{\tilde{\sigma}_i},
\end{equation}
where $\tilde{\mu}_i$ and $\tilde{\sigma}_i$ represent the median and interquartile range (IQR) of Errt $i$ at timestamp $t$. Subsequently, the max function is applied to aggregate $ano_i(t)$, obtaining the anomaly score at time $t$:
\begin{equation}
ano(t)=\max_i(ano_i(t)).
\end{equation}

Finally, a simple moving average is adopted to generate the final anomaly score. If $ano(t)$ exceeds a fixed threshold, the timescale $t$ is labelled as an anomaly.

\section{Experiments}
In this section, we go through experiments to demonstrate the effectiveness of PGMA on anomaly detection.

\begin{table}[t!]
\centering
\caption{Statistical summary of datasets}
\label{table:summary}
\renewcommand{\arraystretch}{1.2}
\setlength{\tabcolsep}{6pt}
\begin{tabular}{@{}lcccc@{}}
\toprule
\textbf{Datasets} & \textbf{Train} & \textbf{Test} & \textbf{Nodes} & \textbf{Anomalies(\%)} \\ \hline
SWAT & 496800 & 449919 & 51 & 11.98 \\
MSL  & 58317  & 73729  & 55 & 10.72 \\
SMAP & 135183 & 427617 & 25 & 13.13 \\
SMD  & 708377 & 708393 & 38 & 4.16  \\
\bottomrule
\end{tabular}
\end{table}

\begin{table*}[htbp!]
\centering
\caption{Anomaly Detection Performance}
\label{table:performance}
\renewcommand{\arraystretch}{1.2}
\setlength{\tabcolsep}{6pt}
\begin{tabular}{@{}lcccccccccccc@{}}
\toprule
\multirow{2}*{\textbf{Methods}} & \multicolumn{3}{c}{\textbf{MSL}} & \multicolumn{3}{c}{\textbf{SMAP}} & \multicolumn{3}{c}{\textbf{SMD}} & \multicolumn{3}{c}{\textbf{SWAT}} \\ 
\cmidrule(lr){2-4} \cmidrule(lr){5-7} \cmidrule(lr){8-10} \cmidrule(lr){11-13}
& \textbf{F1} & \text{PRE} & \text{REC} & \textbf{F1} & \text{PRE} & \text{REC} & \textbf{F1} & \text{PRE} & \text{REC} & \textbf{F1} & \text{PRE} & \text{REC} \\ \hline
VAE-LSTM    & 0.8630 & 0.8063 & 0.9278 & 0.7274 & 0.5765 & 0.9853 & 0.8331 & 0.7297 & 0.9706 & 0.8183 & 0.8202 & 0.7365 \\
OmniAnomaly & \underline{0.8826} & 0.8539 & 0.9117 & 0.8434 & 0.7416 & 0.9776 & 0.8241 & 0.7593 & 0.9010 & 0.8137 & 0.9803 & 0.6957 \\
GDN         & 0.8783 & 0.7826 & 0.9981 & \underline{0.8517} & 0.7499 & 0.9854 & 0.8769 & 0.8141 & 0.8503 & 0.7750 & 0.8657 & 0.6414 \\
USAD        & 0.8769 & 0.7920 & 0.9823 & 0.6831 & 0.5754 & 0.8402 & 0.8536 & 0.7595 & 0.9743 & 0.7946 & 0.9197 & 0.6992 \\
TimesNet    & 0.8514 & 0.7992 & 0.9110 & 0.7152 & 0.6007 & 0.8837 & 0.8512 & 0.8776 & 0.8263 & 0.8062 & 0.9736 & 0.6879 \\
MTAD-GAT\cite{mtad-gat}    & 0.8631 & 0.8125 & 0.9187 & 0.6890 & 0.7048 & 0.6733 & \underline{0.8914} & 0.8069 & 0.9954 & 0.8301 & 0.9527 & 0.7287 \\ 
FRNet    & 0.8793 & 0.8631 & 0.8957 & 0.7548 & 0.7388 & 0.7713 & 0.8559 & 0.8803 & 0.8327 & \underline{0.8344} & 0.9716 & 0.7309 \\
\hline
PGMA        & \textbf{0.8972} & 0.8280 & 0.9781 & \textbf{0.8623} & 0.7592 & 0.9976 & \textbf{0.9017} & 0.8863 & 0.9821 & \textbf{0.8374} & 0.9871 & 0.7483 \\
\bottomrule
\multicolumn{10}{l}{The F1 is calculated from pre and rec and is the final measure of the model.}
\end{tabular}

\end{table*}







\subsection{General Setting}
\textbf{Datasets.} We use four cyber–physical system datasets (SWAT, SMD, SMAP, and MSL) to validate anomaly detection under complex relationships. Table I summarizes their statistics, including training/testing nodes, total nodes, and anomaly proportions.

\textbf{Evaluation Metrics.} This study focuses on anomaly detection. Hence, PRE, REC and F1 score are adopted to evaluate the effectiveness of the model.


\textbf{Training Details.} We optimize PGMA using the Adam optimizer with hyperparameters selected via grid search, where the learning rate is searched from $\{0.01, 0.005, 0.0025, 0.00125\}$. The embedding dimensions of the graph and temporal convolution modules are set to 64 and 32, respectively, and the output module is a 2-layer MLP. Models are trained for up to 30 epochs with early stopping (patience=10).

\subsection{Comparison Performance and Analysis}

We compare PGMA with state-of-the-art anomaly detection methods on four datasets, using F1-score as the evaluation metric. Competing methods include reconstruction-based models (e.g., OmniAnomaly, VAE-LSTM) that capture hidden data features via VAE or GAN, and forecasting-based models (e.g., GDN, TimesNet) that leverage RNN, LSTM, or Transformer for historical sequence learning. As shown in TABLE II, PGMA outperforms most baselines on all datasets, achieving an average 1.2\% improvement over the best alternatives.

\begin{table}[t!]
\centering
\caption{Results of Ablation Experiment (F1-score)}
\label{table:ablation}
\renewcommand{\arraystretch}{1.2}
\setlength{\tabcolsep}{8pt}
\begin{tabular}{@{}lcccc@{}}
\toprule
\textbf{Model}  & \textbf{MSL}   & \textbf{SMD}   & \textbf{SMAP}  & \textbf{SWAT}   \\ \midrule
w/o STIA        & 0.8624         & 0.8615         & 0.8141         & 0.7863         \\
w/o PGSL        & 0.8401         & 0.8596         & 0.8357         & 0.7922         \\
PGMA            & \textbf{0.8972} & \textbf{0.9017} & \textbf{0.8623} & \textbf{0.8374} \\ \bottomrule
\end{tabular}
\end{table}


\subsection{Ablation Study}



We conduct ablation experiments on four datasets to evaluate each component of PGMA. First, we replace the periodogram with a static graph to assess its effect, then remove the temporal convolution module to test the role of temporal features. As shown in TABLE III, static graphs without periodic factors reduce accuracy, and removing temporal convolution limits historical information, further degrading performance. These results confirm that both the periodogram and temporal convolution are essential to PGMA’s superiority over baselines.

\subsection{Hyperparameter Analysis}

\begin{figure}[t!]
\centerline{\includegraphics[width=7.5cm]{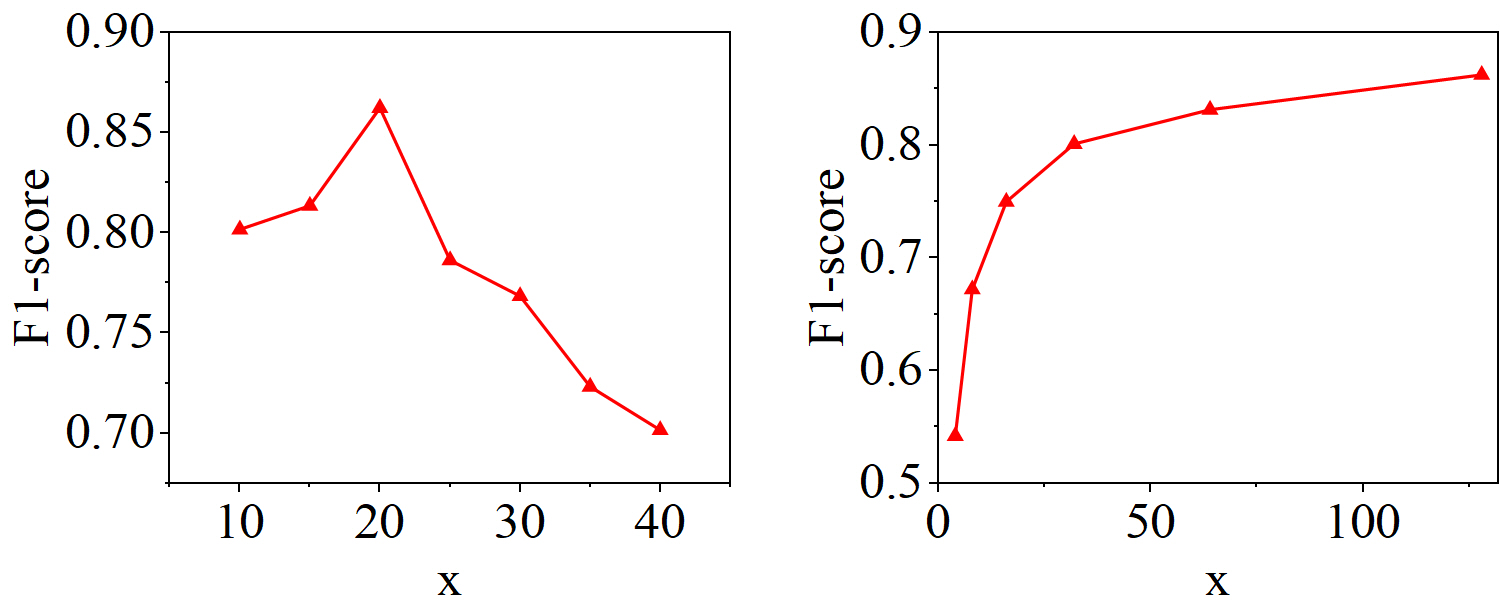}}
\caption{Hyperparameter analysis that Number of neighbors(left) and Number of filters(right).}
\label{fig:Hyper}
\end{figure}


We analyze hyperparameters affecting PGMA’s performance, including the number of neighbors in graph learning (TopK) and the number of filters (output dimension $d^{\prime}$ for temporal and graph convolutions). On the SMAP dataset, we vary $k\in\{10,15,20,25,30,35,40\}$ and $d^{\prime}\in\{4,8,16,32,64,128\}$. As shown in Fig. 4, too many neighbors add noise and increase complexity without improving performance, while more filters enhance model expressiveness and raise the F1-score.





\section{Conclusion}
In this paper, We propose a periodic graph-based multivariate time series anomaly detection method (PGMA) that enhances accuracy and efficiency by integrating periodic graph structure learning with temporal information aggregation. Experiments demonstrate its strong generalization across diverse scenarios, highlighting the effectiveness of periodograms in multivariate analysis. However, PGMA performs less effectively in settings with simple or no spatio-temporal features, and future work will focus on optimizing dynamic graph updates for such cases.


\bibliographystyle{IEEEtran}
\bibliography{arxiv} 

\end{document}